\pdfoutput=1
\def\year{2021}\relax
\documentclass[letterpaper]{article} % DO NOT CHANGE THIS
\usepackage{aaai21}  % DO NOT CHANGE THIS
\usepackage{times}  % DO NOT CHANGE THIS
\usepackage{helvet} % DO NOT CHANGE THIS
\usepackage{courier}  % DO NOT CHANGE THIS
\usepackage[hyphens]{url}  % DO NOT CHANGE THIS
\usepackage{graphicx} % DO NOT CHANGE THIS
\usepackage{natbib}  % DO NOT CHANGE THIS AND DO NOT ADD ANY OPTIONS TO IT
\usepackage{caption} % DO NOT CHANGE THIS AND DO NOT ADD ANY OPTIONS TO IT
\usepackage[bookmarks=false]{hyperref}
\usepackage{subfig}
\usepackage{amsmath}
\usepackage{amsfonts}

\DeclareMathOperator*{\argmin}{arg\,min}
\usepackage{booktabs}
\usepackage{siunitx}
\usepackage{etoolbox}% <-- for bold fonts
\newcommand{\ubold}{\fontseries{b}\selectfont}% <-- for bold fonts
\robustify\ubold% <-- for bold fonts
\usepackage{todonotes}

\usepackage[ruled,linesnumbered]{algorithm2e}

\title{Practical Cross-modal Manifold Alignment for Grounded Language}

\author {
        Andre T. Nguyen,\textsuperscript{\rm 1,2}
        Luke E. Richards,\textsuperscript{\rm 1,2}
        Gaoussou Youssouf Kebe,\textsuperscript{\rm 1}
        Edward Raff,\textsuperscript{\rm 1,2}
        Kasra Darvish,\textsuperscript{\rm 1}
        Frank Ferraro,\textsuperscript{\rm 1}
        Cynthia Matuszek\textsuperscript{\rm 1} \\
}
\affiliations {
    \textsuperscript{\rm 1} University of Maryland, Baltimore County, Baltimore, MD \\
    \textsuperscript{\rm 2} Booz Allen Hamilton, Annapolis Junction, MD \\
    Nguyen\_Andre@bah.com
}

\usepackage{fancyhdr}

\begin{document}
\nocopyright
\maketitle

\begin{abstract}

We propose a cross-modality manifold alignment procedure that leverages triplet loss to jointly learn consistent, multi-modal embeddings of language-based concepts of real-world items. 
Our approach learns these embeddings by sampling triples of anchor, positive, and negative data points from RGB-depth images and their natural language descriptions. 
We show that our approach can benefit from, but does not require, post-processing steps such as Procrustes analysis, in contrast to some of our baselines which require it for reasonable performance. 
We demonstrate the effectiveness of our approach on two datasets commonly used to develop robotic-based grounded language learning systems, where our approach outperforms four baselines, including a state-of-the-art approach, across five evaluation metrics.

\end{abstract}

\section{Introduction: Grounded Language Acquisition Through the Lens of Manifold Alignment}

As robots become advanced and affordable enough to have in our daily lives, work needs to be done to make these machines as intuitive as possible. Language offers an approachable interface. 
However, understanding how natural language can best be grounded to the physical world is still very much an open problem. Combining language and robotics creates unique challenges that much of the current work on grounded language learning has not yet addressed. 

Acquiring grounded language---learning associations between symbols in language and their referents in the physical world---takes many forms~\cite{harnad1990symbol}. With some exceptions~\cite{thomason2016learning}, the majority~\cite{Krishna2016VisualGC,salvador2017learning} of current work focuses on grounding language to RGB images. Due to the availability of large datasets consisting of up to millions of parallel RGB images and language \cite{MarinIEEEImages,Krishna2016VisualGC,Plummer2015Flickr30kEC}, these tasks typically operate with a large pool of data. Large annotated datasets are rare in the field of grounded language for robotics, especially datasets containing depth information in the form of RGB-D. 

This is a complex problem space, and has been demonstrated successfully in domains as varied as soliciting human assistance with tasks~\cite{Knepper2015}, interactive learning~\cite{She2017InteractiveLO}, and understanding complex spatial expressions~\cite{Paul_IJRR_2018}.
Previous work~\cite{Pillai_AAAI_2018,richards2019learning} has made many simplifying assumptions such as using a bag-of-words language model and focusing on using domain-specific visual features for training classifier models. Our approach relaxes these assumptions: we do not assume any particular form of language model nor any specific visual features. 

In particular, we demonstrate how to recast existing but disparate language and vision domain representations into a joint space. We do so by learning a transform of both language and Red Green Blue Depth (RGB-D) sensor data embeddings into a joint space using manifold alignment. This enables the learning of grounded language in a cross-domain manner and provides a bridge between the noisy, multi-domain perceived world of the robotic agent and unconstrained natural language. In particular, we use triplet loss in combination with Procrustes analysis to achieve the alignment of language and vision. Our approach to alignment attains state-of-the-art performance on the language enhanced University of Washington RGB-D Object Dataset \cite{richards2019learning, Lai_ICRA_2011} as well as on the dataset of \citet{Pillai_AAAI_2018}. Importantly, our approach should be able to integrate with existing robot sensors and models with little additional overhead. The primary contribution of this work is the introduction of an easy to implement manifold alignment approach to the grounded language problem for systems where sensor data representations do not live in the same space. We additionally demonstrate generalizability to the unsupervised setting and examine the contribution of Procrustes analysis post-processing.

\section{Related Work}

We treat the language grounding problem as one of manifold alignment---finding a mapping from heterogeneous representations (commonly the case with language and sensor datasets) to a shared structure in latent space \cite{wang2013manifold}. This makes the assumption that there is an underlying manifold that datasets share, obtained by leveraging correspondences between paired data elements. 

\begin{figure}[tb]
\centering
\vspace{0ex}
\includegraphics[width=.85\columnwidth]{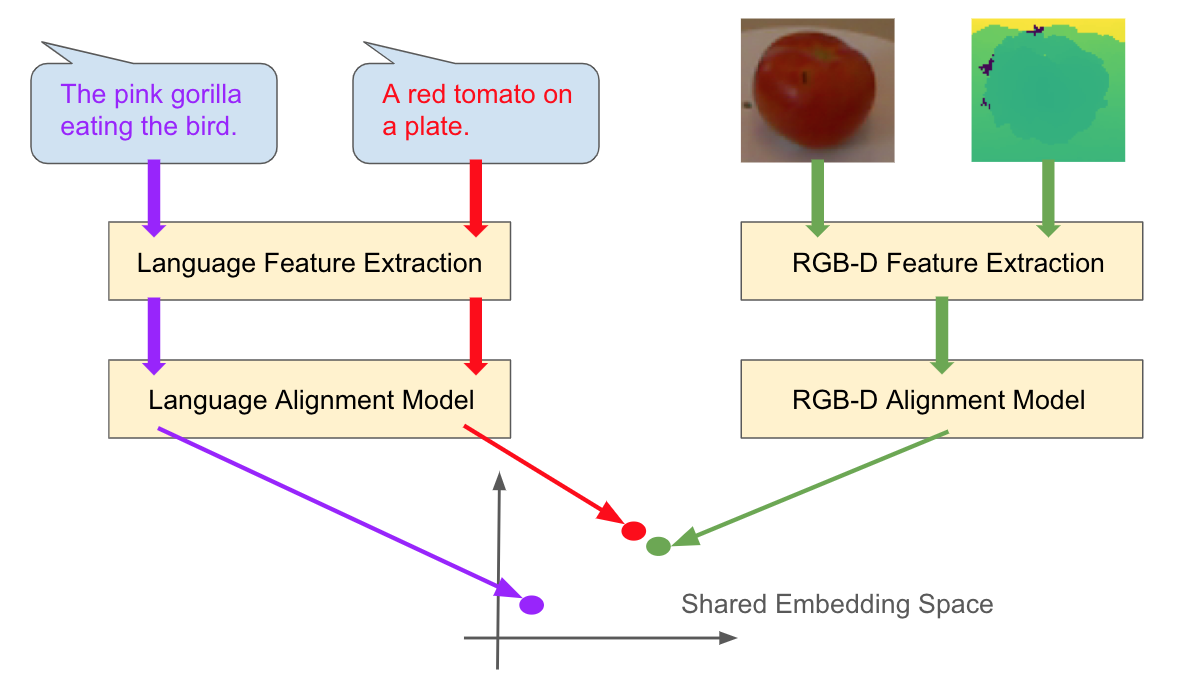}
\vspace{-1ex}
\caption{A language and vision manifold alignment approach to language grounding. On the left side, natural language data is first vectorized by a feature extractor then mapped to a new space using an embedding function. Similarly, on the right side, RGB-D data is vectorized and then embedded to the same space as the language data. The objective is to learn language and vision alignment functions that map similar concepts in the different domains close to each other in the shared embedding space and that map dissimilar concepts far from each other in the shared space. 
}
\label{fig:figs/flowchart.png}
\end{figure}

Jointly learning embeddings for different data domains to a shared latent space can yield a consistent representation of concepts across domains. \autoref{fig:figs/flowchart.png} illustrates the goal of aligning language and vision.

Given $n$ different domains, the manifold alignment task is to find $n$ functions, $f_1,...,f_n$ such that each function maps each $m_i$-dimensional space to a shared latent $M$-dimensional space, $f_i \colon \mathbb{R}^{m_i} \to \mathbb{R}^M  \; , i=1,...,n.$ In our case, $n=2$ where the domains correspond to RGB-D and natural language.

Applying manifold alignment to learning groundings between language and physical context is a relatively novel approach. 

Most prior work in this area
focus on the cooking domain using the much larger Recipe1M dataset containing around one million cooking recipes and eight hundred thousand food
images~\cite{salvador2017learning,carvalho2018cross,Fain2019DividingSoTA}. Our work differs from these previous works as we demonstrate the effectiveness of a manifold alignment approach on much smaller data (our datasets have less than one percent of the number of data points in the Recipe1M dataset). \citet{lazaridou-etal-2014-wampimuk} learn a projection of image-extracted features to an existing and fixed language embedding space.
  
In the robotics domain, \citet{Cohen2019GroundingEigenobjects} combine Bayesian Eigenobjects with a language grounding model that maps natural language phrases and segmented depth images to a shared space. This Bayesian Eigenobjects approach is however evaluated on only three classes of objects. 
Moreover, \citet{Choi2020rss} employ nonparametric regression and deep latent variable modeling to transfer human motion data to humanoid robots. 
Finally, \citet{Lu2019ViLBERT:Tasks} introduce ViLBERT, task-agnostic and transferable joint representations of image content and natural language. Our work differs from ViLBERT as we are not tackling the problem of learning joint embeddings but rather the problem of recasting different existing embeddings into a joint space.

\section{Heterogeneous Domain Alignment}

Deep metric learning \cite{Kaya2019DeepSurvey} uses deep neural networks to learn a projection of data to an embedding space where intra-class distances are smaller than inter-class distances. Our intention is that the learned metric and embedding capture the semantics of the paired data. The triplet loss directly encodes the desire that data from a common class be `closer together' than data from other classes \cite{balntas2016learning, schroff2015facenet}. In particular, triplet loss seeks to minimize the distance between an anchor point and a positive point belonging to the same class as the anchor, while maximizing the distance between the anchor point and a negative point belonging to a different class. Given an anchor $x_a$, positive $x_p$, and negative $x_n$ triplet each in $\mathbb{R}^m$, we seek to minimize the following triplet loss where $d$ is a distance metric, $f$ is the embedding function we want to learn, and $\alpha$ is a margin enforced between positive and negative data pairs:

\begin{equation} \label{eq:triplet}
L=\max\left(d\left(f(x_a),f(x_p)\right) - d\left(f(x_a),f(x_n)\right) + \alpha, 0\right)
\end{equation}

Previous work has used triplet loss for learning metric embeddings, for example \citet{hermans2017defense} maps similar data from homogeneous domains closer to each other in a shared lower-dimensional latent space. Our approach, in contrast, is to use data from heterogeneous domains to learn the metric embeddings based on triplet loss.

More specifically, we wish to learn two embedding alignment functions $f_v$ and $f_l$ that map RGB-D (i.e., ``vision'' $f_v$) and language ($f_l$) data respectively to a shared representation space. In order to jointly learn $f_v$ and $f_l$, we use the triplet loss but select triplets to be cross-domain. In particular, we select triplets such that the anchor, the positive, and the negative can independently belong to either domain. For example, in the case where the anchor and negative come from the vision domain and the positive comes from the language domain, the loss for that triplet is:

\begin{equation} \label{eq:triplet2}
L=\max\left(d(f_v(x_a),f_l(x_p)) - d(f_v(x_a),f_v(x_n)) + \alpha, 0\right)
\end{equation}

\noindent In the above example, $x_a$ could be a cat RGB-D image, $x_p$ a textual description of a cat, and $x_n$ a toaster image. Our primary method, called ``Triplet Method'' throughout this paper, uses cosine distance as the distance metric $d$.

Once the embedding alignment transformations $f_v$ and $f_l$ are learned, an optional fine-tuning step can be included in the form of a Procrustes analysis \cite{gower1975generalized} which finds the optimal translation, scaling, and rotation of two shapes to minimize the Procrustes distance between the shapes. The Procrustes distance is the Euclidean distance between the shapes after the learned optimal translation, scaling, and rotation of shapes. An optimal rotation matrix $R$ is found such that the Euclidean distance between the shapes after translation and scaling is minimized

\begin{equation} \label{eq:rotation}
R^{*} = \argmin \left| \left| \frac{f_v(X_v) - m_v}{\|f_v(X_v) - m_v\|_F} - \frac{ f_l(X_l) - m_l}{\| f_l(X_l) - m_l\|_F} R^T\right| \right|_2.
\end{equation}

\noindent where $X_v$ and $X_l$ are the vision and language data respectively (where rows from each domain form pairs), where $m_v$ and $m_l$ are the means of $f_v(X_v)$ and $f_l(X_l)$, and $\|\cdot\|_F$ is the Frobenius matrix norm. All the Procrustes analysis parameters are fit using the training set. As we will show, our method can benefit from, but does not require, Procrustes analysis, in contrast to some of our baselines which require it for reasonable performance. The full training procedure for the triplet method is described at a high level in \autoref{alg:overvew}.

\begin{algorithm}[!tb]
\DontPrintSemicolon
\KwIn{Dataset $X$ of paired RGB-D and language feature vectors $(x_v,x_l)$.}
\KwOut{Embedding alignment functions $f_v$ and $f_l$ that map RGB-D and language data to a shared embedding space and a trained Procrustes transform.}

$f_v, f_l \leftarrow $ randomly initialized neural networks with parameters $\theta_v$ and $\theta_l$ respectively\;
\While{not converged}{
    $x_a \gets $ randomly selected vision or language feature vector from $X$\;
    $x_p \gets $ randomly selected vision or language feature vector from $X$ belonging to the same class as $x_a$\;
    $x_n \leftarrow $ randomly select any other vision or language feature vector from $X$ belonging to a different class than $x_a$ and $x_p$\;
    Incur loss $L$ using \autoref{eq:triplet2}, and backpropogate to update parameters $\theta_v$ and $\theta_l$ \;
}

\tcp{Compute Procrustes parameters.}
$m_v \leftarrow \frac{1}{|X|}\sum_{\forall (x_v, x_l) \in X} f_v(x_v)$ \;
$m_l \leftarrow \frac{1}{|X|}\sum_{\forall (x_v, x_l) \in X} f_l(x_l)$ \;
$s_v \leftarrow \|f_v(X_v) - m_v\|_F$ \;
$s_l \leftarrow \|f_l(X_l) - m_l\|_F$ \;
$R \leftarrow $ solution to \autoref{eq:rotation}\;

\tcp{Return the aligned embedding functions and Procrustes parameters.}
\KwRet{$f_v, f_l, m_v, m_l, s_v, s_l, R$}

\caption{Training Procedure for Triplet Method}
\label{alg:overvew}
\end{algorithm}

\section{Experiments}
\subsection{Grounded Language Data and Evaluation}

\begin{figure}[!]
\centering
\captionsetup[subfigure]{labelformat=empty}
\subfloat[]{\includegraphics[width=.25\columnwidth]{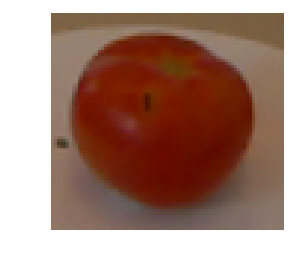}}\qquad
\subfloat[]{\includegraphics[width=.25\columnwidth]{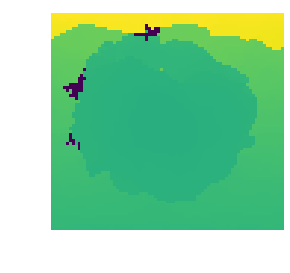}}\qquad
\subfloat[]{\shortstack[l]{\footnotesize A fruit that is\\ \footnotesize round and red\\ \footnotesize and best with\\ \footnotesize salads and\\ \footnotesize sandwiches. \bigskip}}\\
\vspace{-3\baselineskip}
\subfloat[]{\includegraphics[width=.25\columnwidth]{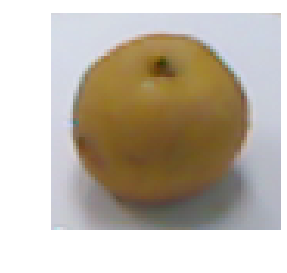}}\qquad
\subfloat[]{\includegraphics[width=.25\columnwidth]{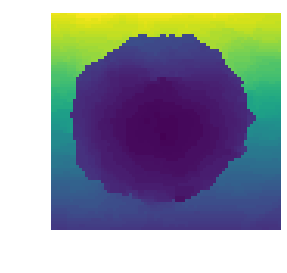}}\qquad
\subfloat[]{\shortstack[l]{\footnotesize This is a \\ \footnotesize piece of fruit.\medskip\medskip \bigskip}}\\
\vspace{-3\baselineskip}

\subfloat[]{\includegraphics[width=.25\columnwidth]{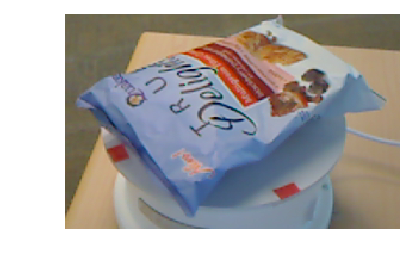}}\qquad
\subfloat[]{\includegraphics[width=.25\columnwidth]{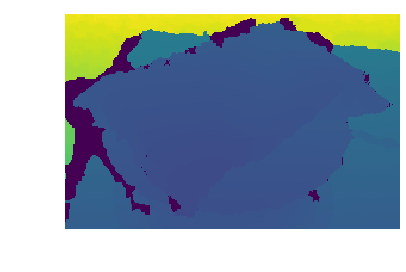}}\qquad
\subfloat[]{\shortstack[l]{ \footnotesize A bag of \\ \footnotesize frozen fruit. \bigskip}}\\
\vspace{-2\baselineskip}

\caption{Example data from the Tomato, Pear, and Food Bag classes. Each row corresponds to a class, and columns correspond to RGB, depth, and text descriptions.}
\label{fig:figs/data}
\end{figure}

We use the same RGB-D object dataset used by \citet{richards2019learning}, which extends the classic and well-known University of Washington object dataset~\cite{Lai_ICRA_2011} with natural language text descriptions. The dataset consists of 7,455 RGB-D image and text description pairs where the pairs each belong to one of 51 classes and where the number of data points per class range from 33 to 366. \autoref{fig:figs/data} shows example data from the Tomato, Pear, and Food Bag classes. The three examples shown in \autoref{fig:figs/data} illustrate how ambiguity can occur in natural language, as all three classes can be described using the word ``fruit.'' During evaluation, we desire that a good approach map the RGB-D and language representations of each object class near each other but far from the representations of objects from other classes.

\subsection{Models}

For the language feature extraction model $b_l$, we use a 12-layer BERT model pre-trained on lowercase English text \cite{Devlin2018BERT:Understanding}. The resulting BERT embedding is 3,072 dimensional.

For the vision feature extraction $b_v$, we use a ResNet152 pre-trained on ImageNet \cite{HeDeepRecognition} with its last fully connected layer removed. The depth component is dealt with via colorization (which we shall call D2RGB) in a similar manner to the procedure from \citet{Eitel2015MultimodalDL} which encodes a depth image as an RGB image where the information contained in the depth data is spread across all three RGB channels. This allows us to use the same pre-trained ResNet to process both the RGB image and the transformed depth information. The vectors resulting from the RGB images and the D2RGB depth-to-RGB colorization are concatenated to create a final 4,096 dimensional RGB-D vision embedding. This gives us $b_v(x_{RGB-D}) = [\text{ResNet}(x_{RGB}); \text{ResNet}(\text{D2RGB}(x_{D}))]$.

\citet{Lu2019ViLBERT:Tasks} introduce ViLBERT, joint representations of images and natural language. We note that while a pre-trained ViLBERT embedding could be used for the vision and language feature extraction, we do not use ViLBERT as our feature extractor in our experiments.

This is because ViLBERT learns vision and language embeddings jointly, and so the representations are already designed to work together. Our interest is in adapting embedding that have no prior relation, since it is increasingly common to use pre-trained models for different sub tasks.

In our experiments, the network architectures for our alignment models consist of an input layer, two hidden layers of size equal to the input layer size, and an output layer that has the size of the desired embedding dimensionality, set to 1,024 in our experiments. Rectified linear units were used as hidden layer activation functions, and Adam was used as the optimizer \cite{Kingma2014Adam:Optimization}. The triplet loss uses cosine distance as the distance metric with a margin of $\alpha = 0.4$, where we did not tune the margin. The embedding space is chosen to be 1,024-dimensional and we fix the pre-trained feature extraction models $b_l$ and $b_v$ during training, only optimizing the alignment models.

The fixing of the feature extraction models directly connects to the robotics use-case where feature extraction model outputs may be used for multiple tasks and where there may be memory and latency constraints. By not having to store and process data through multiple feature extraction models, our approach is advantageous in how it can fit on top of existing state-of-the-art algorithms used by the robot for separate tasks. To illustrate, the feature extraction models together have \num{167626048} parameters, and the alignment models together have \num{59785216} parameters. In the case of an existing system with language and vision models currently being used for other tasks, the integration of manifold alignment would result in a $36\%$ increase in the number of parameters if the feature extraction models are reused whereas an $136\%$ increase in the number of parameters would occur if the feature extraction models are retrained. 

\subsection{Baselines}
We compare our manifold alignment method with the following baselines. We also augment each of these baselines with a Procrustes analysis for additional, stronger baselines.

\subsubsection{Canonical Correlation Analysis}
Canonical Correlation Analysis (CCA) finds the linear combinations of variables within each of two datasets that maximizes the linear correlation between the combinations from each of the datasets \cite{hotelling1992relations}. 

\subsubsection{Deep CCA}
Deep Canonical Correlation Analysis (Deep CCA) is an extension of CCA where a nonlinear transformation of two datasets is learned to maximize the post-transformation linear correlation \cite{andrew2013deep}. 
Deep CCA suffers from known numerical stability issues due to the need to backpropagate through eigen-decompositions. Additionally, mini-batched stochastic gradient descent cannot be directly used for optimization as correlation is a function of the training data in its entirety and does not decompose into a sum over data points. As a result, care needs to be taken and additional tricks potentially used when training Deep CCA \cite{WangBackpropagation-FriendlyEigendecomposition, Wang2015StochasticIterations}. In particular, we found it necessary to reduce the dimensionality of the embedding space from 1,024 to 64 for Deep CCA in order to avoid numerical instability during the training process. Testing with larger dimensions resulted in a failure to converge. 

\subsection{Manifold Metrics}

To evaluate the quality of the manifolds learned, we will use the three metrics specified below: Mean Reciprocal Rank (a measure of global order preservation), K-Nearest Neighbors (a measure of local order preservation), and Distance Correlation (a measure of global absolute distance preservation). A successfully manifold alignment approach should perform well in all three of these tasks. We do not argue that these are sufficient for determining all aspects about a manifold's quality, but posit that they are useful to the tasks we are concerned with.

\subsubsection{Mean Reciprocal Rank}

Given an image and text pair, we can compute the distance in the joint embedding space between the text element and all data points in the vision domain. These distances can then be ranked with $1$ being the closest, $2$ being the second closest, and so forth. Common in information retrieval, Mean Reciprocal Rank (MRR) is the average across the data of the multiplicative inverse of the rank in embedding space of the nearest item from the same class that comes from the other domain \cite{Craswell2009MeanRank}.

\subsubsection{Distance Correlation}

Intuitively, if two embedding manifolds are aligned, distances in one embedding should be correlated to distances in the other embedding. Specifically, if we select two image and text pairs, the distance between them in the vision embedding should be correlated with the distance between them in the language embedding space. To capture this property, we randomly select \num{10000} pairs of image and text pairs and compute the distance between them. The Pearson correlation is then computed between the vision space distances and the language space distances, resulting in a metric between $-1$ and $1$ where closer to $1$ means better alignment. We call this metric Distance Correlation (DC) in this paper. The sampling is done due to the prohibitive cost to compute the pairwise correlation for the entire dataset. 

\subsubsection{K-Nearest Neighbors}
As a final metric, we use K-Nearest Neighbors (KNN) classification accuracy with $K=5$ in our experiments.
This metric captures what performance would look like in an applied setting where a robot may need to associate natural language with a visual concept.

\section{Supervised Alignment Evaluation}

\subsection{Grounded Language Learning} \label{sec:ground_lang_results}

Our ultimate goal for manifold alignment is to enable the grounding of language to referents in the physical world. To directly assess the effectiveness of cross-modal manifold alignment for grounded language, we evaluate the aligned embeddings on the task of determining which objects in RGB-D space correspond to a given language description. In particular, every text description datum can be considered a separate classification task where the goal is the binary classification of all RGB-D images as relevant or not relevant given the text description. 

\begin{figure}[!tb]
\centering
\vspace{0ex}
\includegraphics[width=180pt]{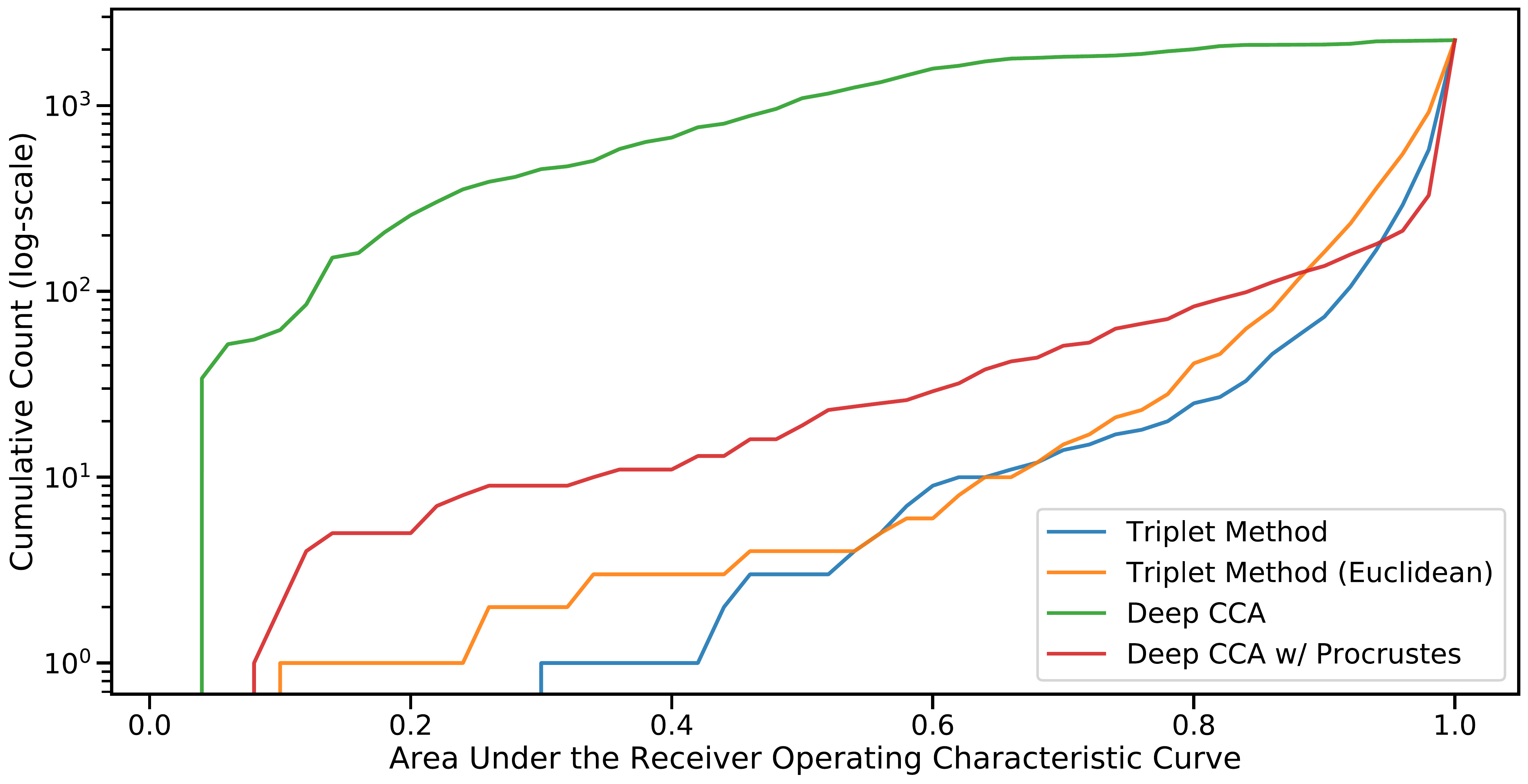}
\vspace{-1ex}
\caption{Grounded language task cumulative counts over AUCs. Lower is better and perfect classification performance lies in bottom-right corner. Every text description datum can be considered a separate classification task (with its own AUC) where the goal is the binary classification of all RGB-D images as relevant or not relevant given the text description. The x-axis represents AUC score values, and the y-axis represents the number of classification tasks with an AUC less than a particular value.  For any particular AUC score, our triplet method has more better scoring tasks than Deep CCA.}
\label{fig:better_auc}
\end{figure}

For each of the classification tasks, an Area Under the Receiver Operating Characteristic Curve (AUC) score is obtained. \autoref{fig:better_auc} shows cumulative counts over AUCs. We note that for any particular AUC score, our triplet method has more better scoring tasks than Deep CCA. In other words, Deep CCA has more and worse failure cases. We also compare our triplet method which uses cosine distance with a version of our triplet method that uses Euclidean distance instead. 
This ablation finds our cosine method best.

\autoref{tab:gl_metrics} summarizes the mean micro and macro averaged F1 scores across methods. 
The triplet method outperforms all of the other methods on the grounded language task. For the computation of F1 scores, the distance between the text description element and RGB-D image element in the shared space was computed for each datum pair in the training set. The relevance distance threshold was set to the mean of these distances plus a standard deviation. 

A comparison of the achieved mean macro averaged F1 score of $0.725$ for the triplet method in the known class scenario with the $0.689$ macro averaged F1 score reported in \citet{richards2019learning}, the current state-of-the art on this dataset, shows a $5.2\%$ improvement and suggests that a manifold alignment approach to grounded language is promising, attaining at least similar or likely better performance than traditional word-as-classifier models. The triplet method without Procrustes achieves a higher $9.9\%$ improvement in macro averaged F1 score, but we will later discuss our preference for the triplet method with Procrustes.

\begin{table}[!tb]
\small
\centering
\begin{tabular}{@{}l l l@{}}
\toprule
 Algorithm & {Avg Micro F1} & {Avg Macro F1} \\ 
  \midrule
Triplet Method   & \textbf{0.983} & 0.725    \\ 
Trip. Met. (w/out Procrustes)  & 0.978 & \textbf{0.757}  \\ 
Trip. Met. (Euclidean)  & 0.969 & 0.727    \\ 
Trip. Met. (Eucl. w/out Proc.)  & 0.952 & 0.714    \\ 

Trip. Met. (unsupervised)  & 0.963 & 0.698  \\ 
Trip. Met. (unsup. w/out Proc.)  &  0.941 & 0.685 \\ 

Cosine Baseline (w/out Proc.)   &  0.542 & 0.337  \\
Cosine Baseline (w/ Proc.)   & 0.441 & 0.318 \\
CCA  & 0.455 & 0.331   \\ 
CCA (w/ Procrustes)  & 0.567 & 0.294   \\ 

Deep CCA  & 0.0261 & 0.0246   \\ 
Deep CCA (w/ Procrustes)  & 0.855 & 0.716   \\ 
\bottomrule
\end{tabular}
\caption{Metrics for grounded language task evaluated on held out test set. Best results are \textbf{bolded}.}
\label{tab:gl_metrics}
\end{table}

\subsection{Effective Deep Metric Learning using Triplet Loss}

\begin{table}[!tb]
\small
\centering
\begin{tabular}{@{}l l l l@{}}
\toprule
Algorithm              & {MRR}          & {KNN}          & {DC}             \\ \midrule
Triplet Method        & 0.802 & 0.787 & 0.686  \\
Trip. Met. (w/out Procrustes)  & 0.758 & 0.742 & 0.692   \\
Trip. Met. (Euclidean)        & 0.724 & 0.702 & 0.693  \\
Trip. Met. (Eucl. w/out Proc.)  &  0.673 & 0.648 & 0.685   \\

Trip. Met. (unsupervised)  & 0.754 & 0.736 & \textbf{0.773}  \\ 
Trip. Met. (unsup. w/out Proc.)  & 0.688 & 0.665 & 0.725  \\ 

Cosine Baseline (w/out Proc.)   &  0.208 & 0.181 & 0.0306  \\
Cosine Baseline (w/ Proc.)   & 0.113 & 0.0965 & -0.00108  \\

CCA                    & 0.0353 & 0.0267 & 0.0400         \\
CCA (w/ Procrustes)      & 0.144 & 0.122 & 0.0665          \\

Deep CCA               & 0.0232 & 0.0116 & 0.377          \\
Deep CCA (w/ Procrustes) & \textbf{0.870} & \textbf{0.860} & 0.359          \\ \bottomrule
\end{tabular}
\caption{Evaluation of manifolds using Mean Reciprocal Rank (MRR), K-Nearest Neighbors (KNN), and Distance Correlation (DC) as metrics. Higher is better for all metrics.}
\label{tab:manifold_metrics}
\end{table}

\autoref{tab:manifold_metrics} shows the MRR, KNN accuracy, and DC for the triplet method as well as for our baselines.
We find that while the triplet method has the highest DC and strong MRR and KNN accuracy, providing consistent performance across all manifold metrics, Deep CCA with the addition of Procrustes analysis has the highest MRR and KNN, at the cost of a $1.9\times$ lower DC compared to our new approach. This disparity in performance means that Deep CCA with Procrustes is not learning a holistically useful manifold. As we saw in \autoref{sec:ground_lang_results}, this translates to worse performance for grounded language learning.  Deep CCA without Procrustes has a significantly reduced, and in fact the worst, MRR and KNN accuracy. CCA with and without Procrustes analysis both have poor performance. These results demonstrate the value of using Procrustes to improve the quality of a manifold alignment at little effort. We also note that while Procrustes is crucial for CCA and Deep CCA, our triplet method remains strong with only a slight decrease in MRR and KNN accuracy when Procrustes analysis is ablated.

\begin{figure}[!h]
\centering
\vspace{0ex}
\includegraphics[width=200pt]{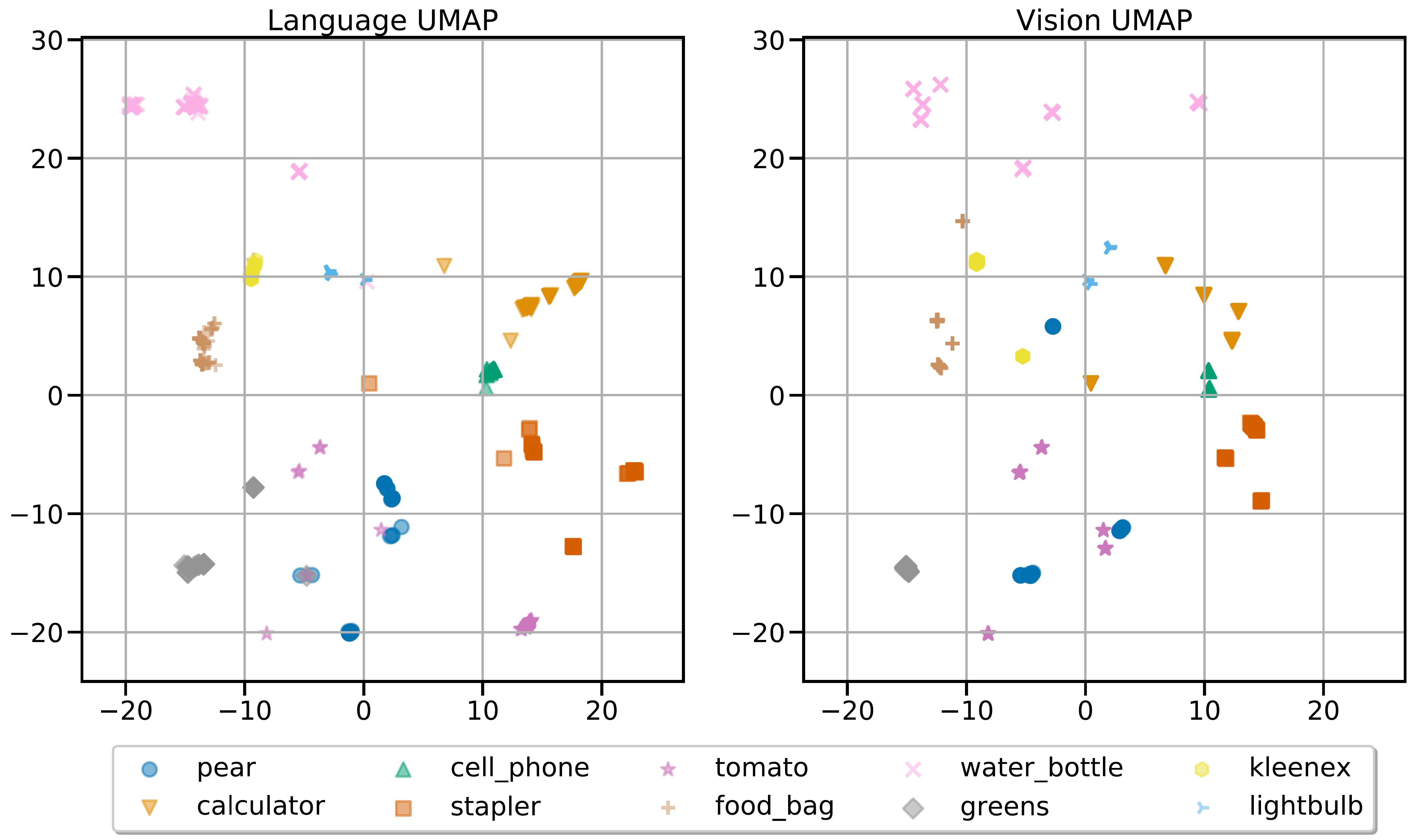}
\vspace{-1ex}
\caption{Test set UMAP of the Triplet Method. 10 randomly selected classes are plotted.}
\label{fig:umap_triplet}
\end{figure}

\begin{figure}[!h]
\centering
\vspace{0ex}
\includegraphics[width=200pt]{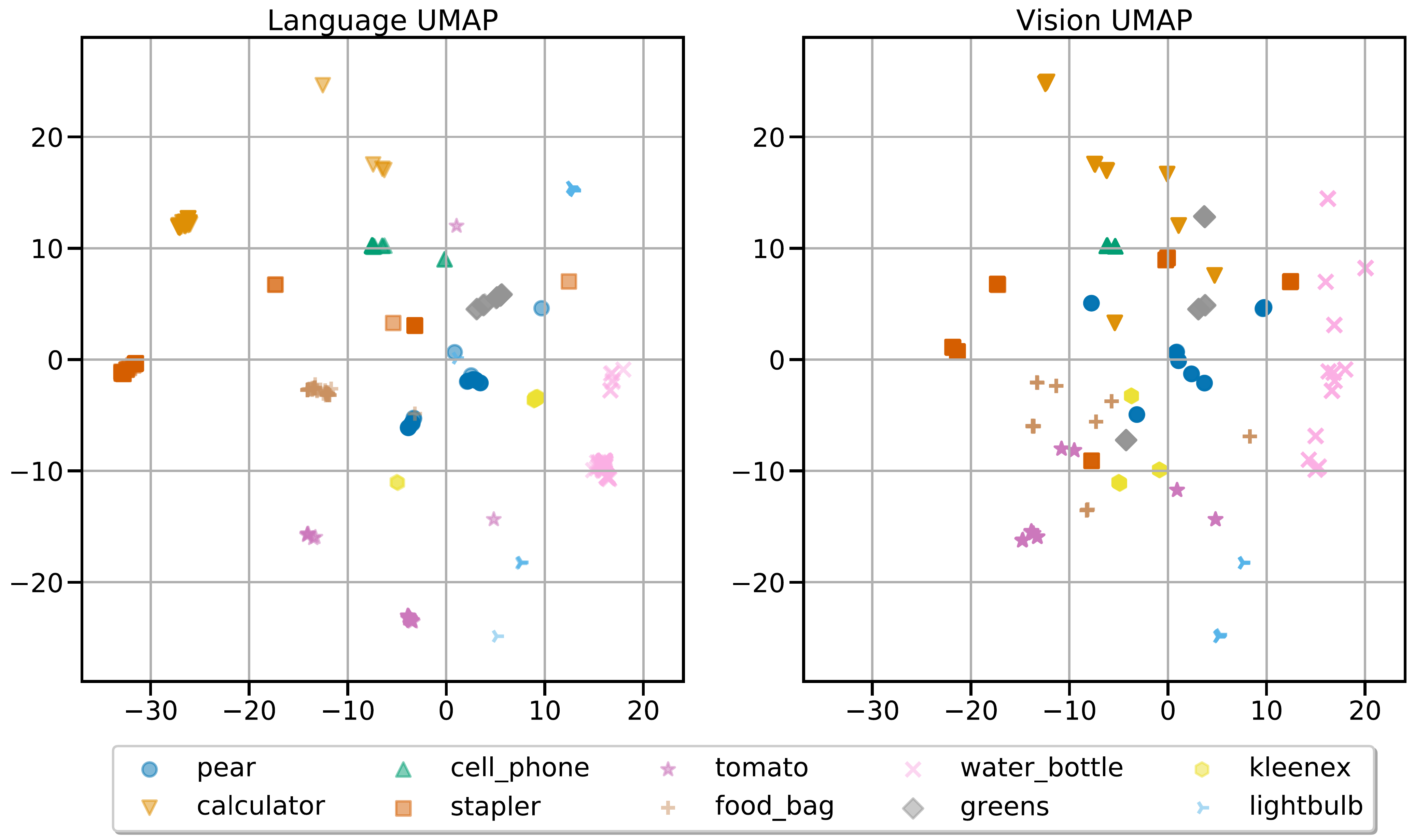}
\vspace{-1ex}
\caption{Test set UMAP of Deep CCA with Procrustes. 10 randomly selected classes are plotted.}
\label{fig:umap_dcca}
\end{figure}

To help confirm that our approach learns good manifolds, we would expect a visualization of the vision and language domains to have similar structure. We do this using UMAP \cite{McInnes2018UMAP:Reduction}, which preserves global structure. 
\autoref{fig:umap_triplet} shows the UMAP for the triplet method. Ten randomly selected classes are plotted for legibility purposes. We observe that classes are generally well clustered (items are close to other items from the same class and classes are separated) and are projected to similar locations across both the language and vision domains. Note that using our new approach, classes with wide dispersion (e.g., the water bottle) or compactness (e.g., cell phone) share this structure across domains. 
\autoref{fig:umap_dcca} shows the UMAP for Deep CCA with Procrustes. In contrast with the triplet method, we observe that while data is well clustered in the language domain, data is less well clustered in the vision domain, in particular when it comes to class separation. Class alignment across domains is also less evident. While classes such as ``cell phone'' and ``food bag'' are well aligned, other classes such as ``kleenex'' and ``calculator'' are not. In these cases the structure is not successfully shared between the domains, indicating a lesser quality as a manifold.

\begin{figure}[!tb]
\centering

\subfloat[Triplet Method]{\includegraphics[width=.5\columnwidth]{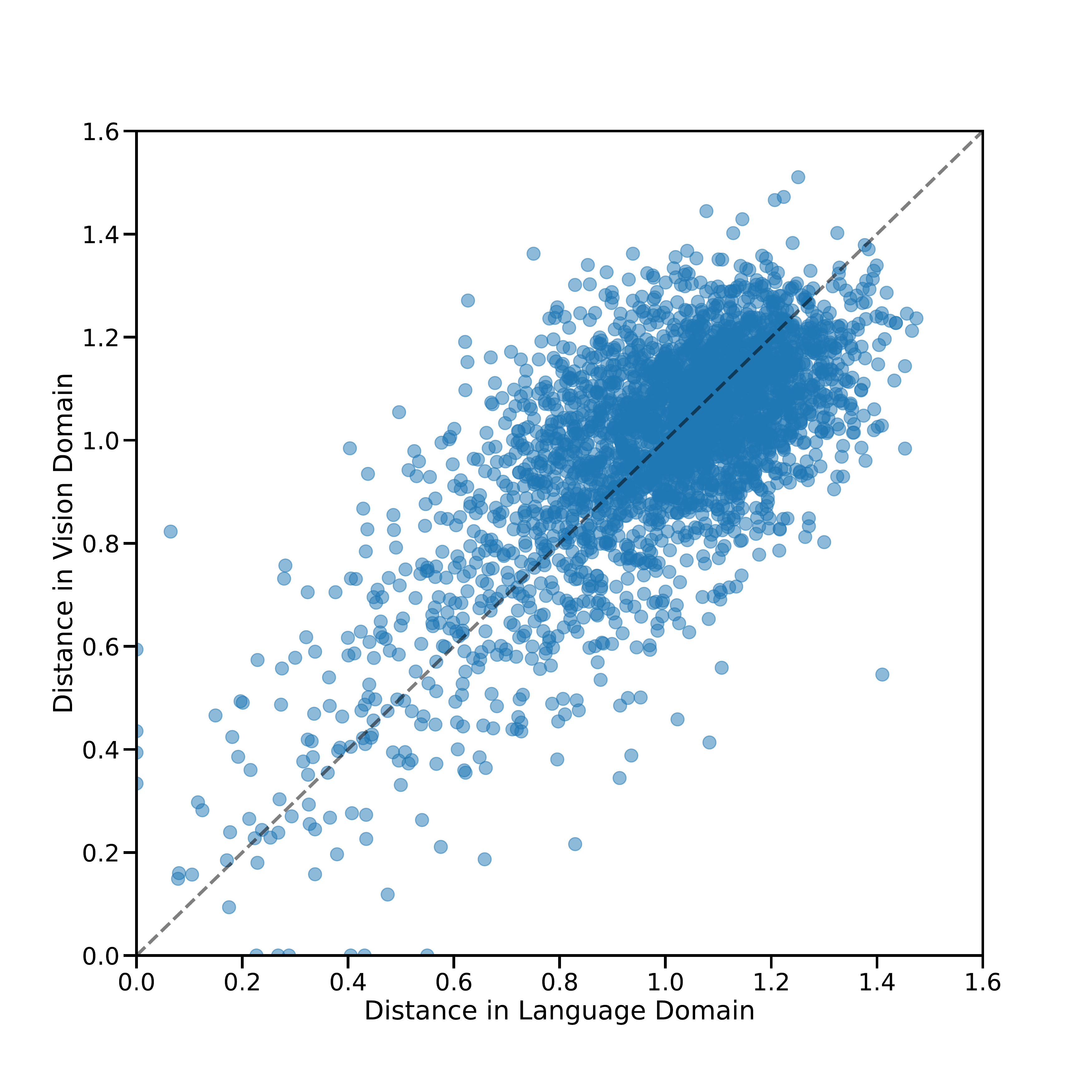}}
\subfloat[Deep CCA with Procrustes]{\includegraphics[width=.5\columnwidth]{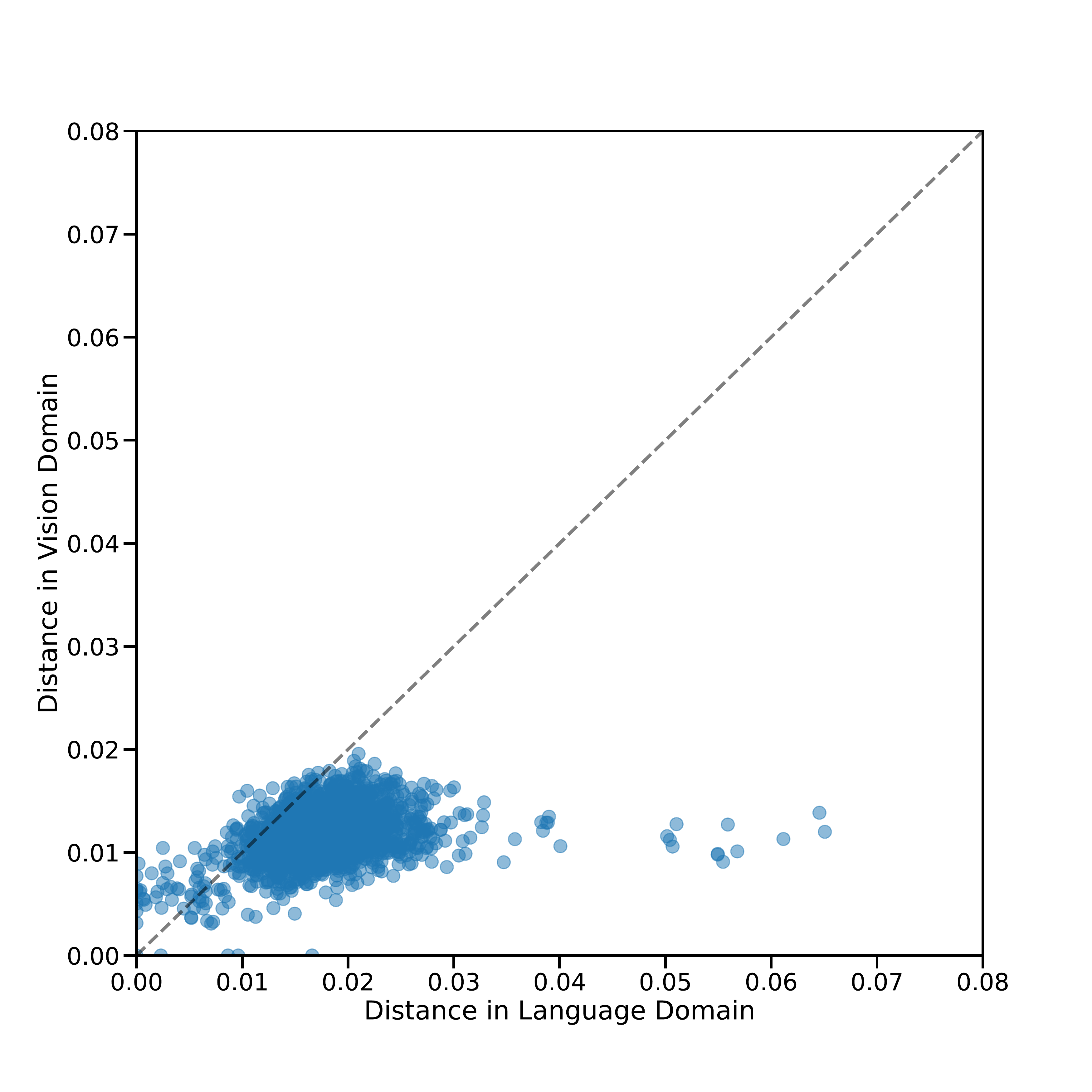}}\\

\caption{Distance Correlation visualization for the Triplet Method and for Deep CCA with Procrustes. Pairs of image and text pairs are randomly selected and the distance between them is plotted, with the x-axis representing the distance in the language domain and the y-axis representing the distance in the vision domain. The dashed line represents where points should lie under perfect manifold alignment.}
\label{fig:figs/corrs}
\end{figure}

Additionally, we can gain more insight into the DC results by plotting the vision space distances and the language space distances to compare their relationships. Subplots (a) and (b) in \autoref{fig:figs/corrs} respectively show the distance relationships for the triplet method and Deep CCA with Procrustes. While the triplet method has the desired linear relationship between distances, Deep CCA with Procrustes lacks the desired relationship that would indicate well aligned manifolds.

\subsection{Understanding the Contribution of Procrustes Analysis and Triplets}  

To better understand the role played by Procrustes analysis, we run ablation experiments, separately removing each of the Procrustes analysis components (translation, scaling, and rotation) one by one. 

\autoref{tab:ablation_triplet_cosine} shows metrics for Procrustes analysis ablations on the triplet method. Metrics stay relatively similar when translation or scaling are removed. When rotation is removed, a decrease in MRR and KNN accuracy is observed without a decrease in DC. 

\autoref{tab:ablation_triplet_euclidean} shows metrics for Procrustes analysis ablations on a variant of the triplet method that uses Euclidean distance instead of cosine distance. Metrics stay similar when translation or rotation are removed. When scaling is removed, a significant decrease in MRR and KNN accuracy is observed. This suggests that the Euclidean distance without Procrustes maps data in each domain to similarly shaped manifolds of different scales. This result is consistent with the formulation of the Euclidean triplet loss, as differently scaled but otherwise similar manifolds can satisfy the relative distance constraints encouraged by the Euclidean triplet loss. This result demonstrates an advantage of the use of cosine distance in this context. A comparison of the performance of the triplet method with its Euclidean variant in \autoref{tab:gl_metrics}, \autoref{tab:manifold_metrics}, and \autoref{fig:better_auc} confirms this advantage.

\begin{table}[!tb]
\small
\centering
\begin{tabular}{@{}l l l l@{}}
\toprule
 Algorithm & {MRR} & {KNN}  & {DC} \\ 
  \midrule
Triplet Method  & 0.802 & 0.787 & 0.686   \\ 

No Translation  & \textbf{0.806} & \textbf{0.790} & 0.679   \\ 

No Scaling  & 0.801 & 0.786 & \textbf{0.696}  \\ 

No Rotation  & 0.750 & 0.733 & 0.693 \\ 
\bottomrule
\end{tabular}
\caption{Ablation metrics where various components of Procrustes analysis are disabled for the Triplet Method.}
\label{tab:ablation_triplet_cosine}
\end{table}

\begin{table}[!tb]
\small
\centering
\begin{tabular}{@{}l l l l@{}}
\toprule
 Algorithm & {MRR} & {KNN}  & {DC} \\ 
  \midrule
Triplet Method  & 0.724 & 0.702 & \textbf{0.693}   \\ 

No Translation  & \textbf{0.729} & \textbf{0.707} & 0.688   \\ 

No Scaling  & 0.0454 & 0.0391 & 0.68   \\ 

No Rotation  & 0.680 & 0.658 & 0.684   \\ 
\bottomrule
\end{tabular}
\caption{Ablation metrics where various components of Procrustes analysis are disabled for the Triplet Method with Euclidean distance.}
\label{tab:ablation_triplet_euclidean}
\end{table}

Similar ablation experiments can be run for Deep CCA with Procrustes analysis. \autoref{tab:ablation_deepcca} suggests that both rotation and scaling are needed for Deep CCA to achieve high MRR and KNN accuracy. 

\begin{table}[!tb]
\small
\centering
\begin{tabular}{@{}l l l l@{}}
\toprule
 Algorithm & {MRR} & {KNN}  & {DC} \\ 
  \midrule
Deep CCA w/ Procrustes  & 0.870 & 0.860 & 0.359    \\

No Translation  & \textbf{0.871} & \textbf{0.862} & 0.363   \\

No Scaling  & 0.0206 & 0.0111 & \textbf{0.378}   \\

No Rotation  & 0.0341 & 0.0214 & 0.352   \\ 
\bottomrule
\end{tabular}
\caption{Ablation metrics where various components of Procrustes analysis are disabled for Deep CCA.}
\label{tab:ablation_deepcca}
\end{table}

We also explore the contribution of using triplets by adding a baseline which seeks to simply minimize the cosine distance between the positive and anchor points in the shared space. \autoref{tab:gl_metrics} and \autoref{tab:manifold_metrics} show the performance of the cosine distance baseline, with and without Procrustes analysis. Overall, the triplet method performs significantly better than the cosine distance baselines. We note that our cosine baseline is similar to the approach taken by \citet{nguyen2020rss}.

\subsection{Comparison of Language Embeddings}

Next, we investigate the effect of better feature extraction. Sentence-BERT (SBERT) is a sentence embedding oriented modification of BERT that achieves better performance on Semantic Textual Similarity (STS) tasks \cite{reimers2019sentence}. 
We compare a BERT-based version of our triplet method to an off-the-shelf SBERT version and a fine-tuned SBERT version. We fine-tune SBERT using pairs of object descriptions from the same extended University of Washington dataset. 
Pairs describing the same instance of an object are given a score of 5 while pairs describing different instances of an object are given a score of 2.5, and pairs describing different objects are given a score of 0.

\autoref{tab:sbert_gl_metrics} and \autoref{tab:sbert_manifold_metrics} summarize the comparative performance of the language embeddings on the downstream grounded language task and manifolds. Fine-tuned SBERT leads to the highest quality manifold. This follows intuition and suggests that the use of higher quality original embeddings of sensor data leads to higher quality aligned representations. Note that we retrained the BERT based triplet method for this experiment, hence the slightly different (but nearly identical) metrics when compared to the numbers reported in \autoref{tab:gl_metrics} and \autoref{tab:manifold_metrics}.

\begin{table}[!tb]
\small
\centering
\begin{tabular}{@{}l l l@{}}
\toprule
 Algorithm & {Avg Micro F1} & {Avg Macro F1} \\ 
  \midrule
Triplet Met. (BERT)  & \textbf{0.984} & 0.735    \\ 
Triplet Met. (SBERT)  & 0.982 & \textbf{0.748}  \\ 
Triplet Met. (SBERT fine-tuned)  & \textbf{0.984} & 0.734    \\ 

\bottomrule
\end{tabular}
\caption{Metrics for grounded language task comparing BERT, SBERT, and a fine-tuned SBERT.}
\label{tab:sbert_gl_metrics}
\end{table}

\begin{table}[!tb]
\small
\centering
\begin{tabular}{@{}l l l l@{}}
\toprule
Algorithm              & {MRR}          & {KNN}          & {DC}             \\ \midrule
Triplet Met. (BERT)       & 0.816 & 0.804 & 0.686  \\
Triplet Met. (SBERT)  & 0.745 & 0.731 & 0.678   \\
Triplet Met. (SBERT fine-tuned)        & \textbf{0.834} & \textbf{0.823} & \textbf{0.731}  \\
\bottomrule
\end{tabular}
\caption{Manifold comparison for BERT, SBERT, and a fine-tuned SBERT. Higher is better for all metrics.}
\label{tab:sbert_manifold_metrics}
\end{table}

\section{Generalizability to Other Settings}

We now investigate if our approach generalizes to situations where unsupervised manifold alignment is needed, and to another dataset with more limited data. 

\subsection{Sampling Negative Examples in an Unsupervised Setting}

So far, the training of the triplet method has assumed the availability of class labels for triplet selection. 
However, the triplet method can still be trained when class ground truth is not available using unsupervised negative example selection. In this setting, the triplets are fixed to have a vision anchor and language negatives and positives. The positive is selected to be the anchor's paired text, and the negative example is chosen through a semantic distance based technique similar to that used in \cite{Pillai_AAAI_2018}. In particular, the cosine distances between all natural language descriptions can be computed, and the negative is sampled from the $25\%$ of descriptions furthest away from the positive description. This can be interpreted as aligning vision to the manifold induced by the language embedding. \autoref{tab:gl_metrics} and \autoref{tab:manifold_metrics} summarize the performance of the triplet method in this unsupervised setting. While there is a decrease in MRR and KNN accuracy, DC remains strong and even increases. On the grounded language task, performance also remains strong with only a $2\%$ decrease in average micro F1 and a $4\%$ decrease in average macro F1. 

\subsection{Effectiveness on a Smaller Dataset}
We also test our triplet method on a dataset \cite{Pillai_AAAI_2018} containing fewer classes and fewer instances per class, with a lower computational cost vision extraction method, depth kernel descriptors \cite{Bo_CVPR_2011} and average values for RGB channel values. Prior work \cite{Pillai_AAAI_2018, Pillai2019RoboNLP} used these same visual feature extraction methods with a word-as-classifier model. 
\citet{Pillai2019RoboNLP} combined depth kernel descriptors and averaged RGB channel values. We concatenate the kernel descriptors and the average RGB channel values into a single vision embedding vector. Each vision vector is paired with a natural language description of the object. 
On this dataset, the triplet method with Procrustes achieves a mean macro F1 score of $0.722$, and the triplet method without Procrustes achieves a mean macro F1 score of $0.729$, both of which are better than but still comparable to the reported $0.714$ for the non-category based model from previous works.

\section{Conclusions}

We explored the use of the triplet loss enhanced with Procrustes analysis for manifold alignment in the context of grounded language. Our approach to alignment achieves state-of-the-art performance on two datasets, and integration with existing robot sensors and models would likely have minimal additional overhead. Next steps include the alignment of more than two modalities, integration with a robot system, and evaluation on a wider variety of tasks.

\section*{Ethics Statement}

We have seen the integration of voice-assistant speakers in homes drastically increase in the recent years, and language may grow to become a preferred method for interacting with AI-enabled assistants. As the application of AI grows beyond location-fixed machines to robots that physically interact with our environment, effectively grounding natural language to referents in the physical world is critical. Advances in this space will have broad societal impacts by improving the quality of robotic assistance for elderly and handicapped users and more generally by improving the productivity and quality of life of all users \cite{koceska2019telemedicine}. We hope that this work will help to improve the state of grounded language for robotics. 

More broadly, the manifold alignment approach has the benefit of interpretability, where the robot's knowledge representation across modalities can be investigated and the manifold studied for AI quality assurance. Also, the ability to recast existing but disparate domain data representations into a joint space is useful in applications outside of robotics. For example, in the space of cyber security, an anti-virus solution may use different models for static and dynamic analysis, and fusing information into a joint space would benefit tasks such as the detection of new virus families \cite{raff2020survey}. 

We note that we have not studied the risks posed by the threat of adversarial attacks which could take the form of data poisoning during the learning of the manifold alignment or prediction time evasion attacks \cite{biggio2018wild}. Successful adversarial attacks on robot systems that can physically interact with our environment have the potential to cause significant damage and danger. While we hope that the explicit manifold representation of robot knowledge will help with the development of defenses against adversarial attacks, much work needs to be done in this area.

\bibliography{static-bib}

\end{document}